\pdfoutput=1

\documentclass[11pt,dvipsnames]{article}

\usepackage[final]{acl}

\usepackage{times}
\usepackage{latexsym}

\usepackage[T1]{fontenc}

\usepackage[utf8]{inputenc}

\usepackage{microtype}

\usepackage{inconsolata}

\usepackage{graphicx}

\usepackage{hyperref}
\usepackage{makecell}
\usepackage[linesnumbered,ruled]{algorithm2e}
\usepackage{colortbl}
\usepackage{float}
\usepackage{stfloats}
\usepackage{enumitem}
\usepackage{booktabs}
\usepackage{multirow}
\usepackage{threeparttable}
\usepackage{amssymb}
\usepackage{adjustbox}
\usepackage{lineno}
\usepackage[x11names]{xcolor}
\usepackage{caption}
\usepackage{amsmath}
\usepackage{marvosym}

\newcommand{\chk}{\checkmark}

\title{Towards Mitigating Modality Bias in Vision-Language Models for Temporal Action Localization}

\author{
    Jiaqi Li\textsuperscript{\rm 1}, Guangming Wang\textsuperscript{\rm 2}, Shuntian Zheng\textsuperscript{\rm 1}, Minzhe Ni\textsuperscript{\rm 1}, Xiaoman Lu\textsuperscript{\rm 1}, \\
    {\bf Guanghui Ye\textsuperscript{\rm 3}, Yu Guan\textsuperscript{\rm 1\Letter}\thanks{Yu Guan is the corresponding author.}}\\
\textsuperscript{1} UVLab, Department of Computer Science, University of Warwick \\
\textsuperscript{2} Department of Automation, University of Cambridge \\
\textsuperscript{3} Department of Computer Science, The University of Sheffield \\
\texttt{\{Jiaqi.Li.16, Shuntian.Zheng, Minzhe.Ni, Xiaoman.Lu, Yu.Guan\}@warwick.ac.uk,}\\
\texttt{gw462@cam.ac.uk, gye4@sheffield.ac.uk}\\
}

\begin{document}
\maketitle
\begin{abstract}

Temporal Action Localization (TAL) requires identifying both the boundaries and categories of actions in untrimmed videos. 
While vision-language models (VLMs) offer rich semantics to complement visual evidence, existing approaches tend to overemphasize linguistic priors at the expense of visual performance, leading to a pronounced modality bias.
We propose ActionVLM, a vision-language aggregation framework that systematically mitigates modality bias in TAL. 
Our key insight is to preserve vision as the dominant signal while adaptively exploiting language only when beneficial.
To this end, we introduce 
(i) a debiasing reweighting module that estimates the language advantage—the incremental benefit of language over vision-only predictions—and dynamically reweights language modality accordingly, 
and (ii) a residual aggregation strategy that treats language as a complementary refinement rather than the primary driver.
This combination alleviates modality bias, reduces overconfidence from linguistic priors, and strengthens temporal reasoning.
Experiments on THUMOS14 show that our model outperforms state-of-the-art by up to 3.2\% mAP.
Our code is available at \url{https://github.com/JiaqiLi404/ActionVLM}

\end{abstract}  
\section{Introduction}
\label{sec:intro}

\begin{figure}[t] 
    \centering
    \begin{adjustbox}{width=1\linewidth,center}
\includegraphics{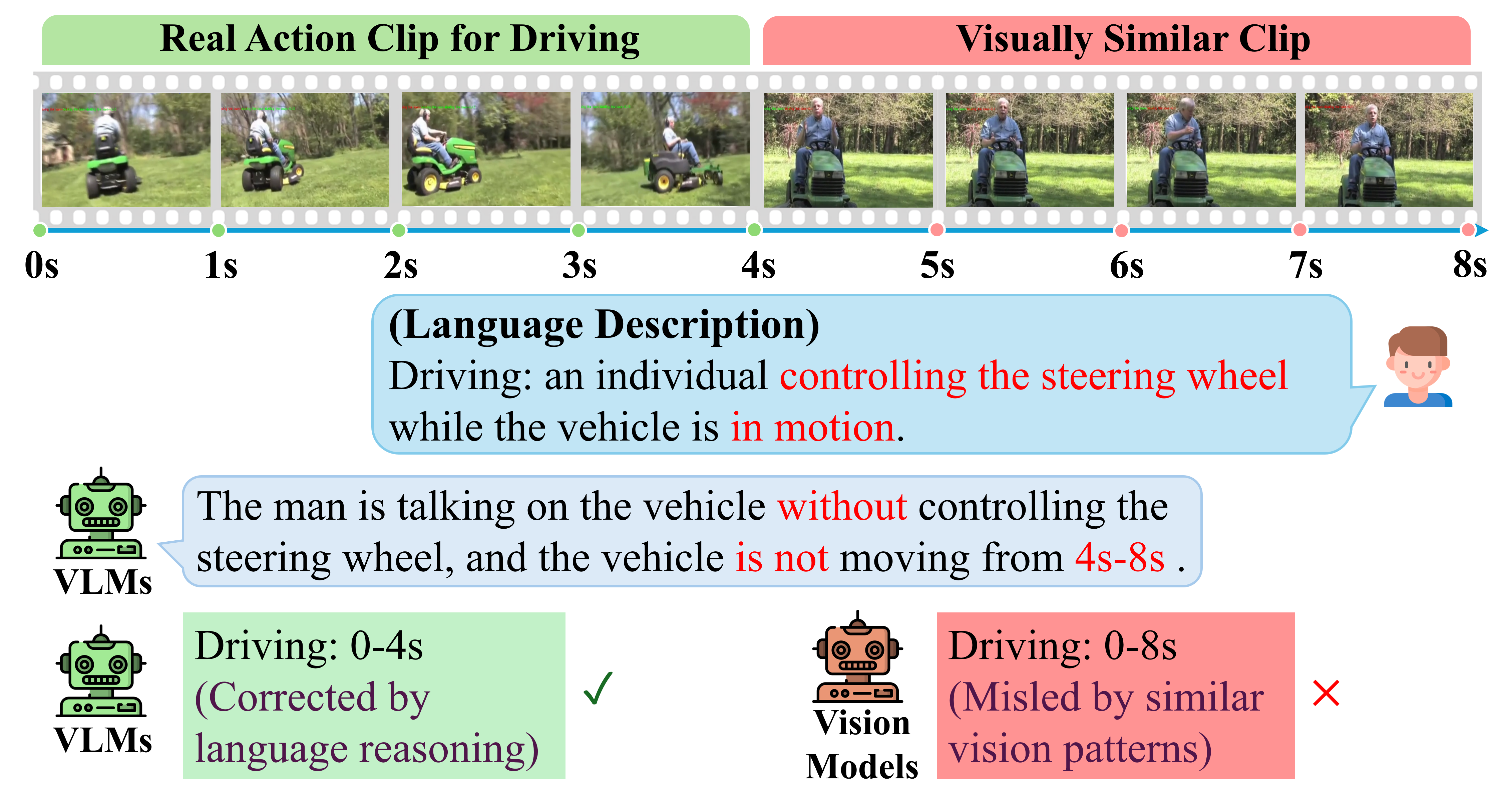} 
\end{adjustbox}
\caption{Visually similar actions are hard to distinguish by pixels alone, but language description provides complementary semantics that resolve the ambiguity. } 
\label{Fig:intension} 
\end{figure}

Temporal Action Localization (TAL) requires identifying the precise start and end of actions in long, untrimmed videos.
This task demands not only robust frame-level recognition but also fine-grained temporal reasoning \cite{wu2024open}.
While purely visual models excel at capturing local appearance and motion patterns, they often struggle when actions share highly similar visual frames \cite{jian2025teaching}.
To alleviate such ambiguities, recent research has turned to Vision-Language Models (VLMs), expecting high-level semantics from language to complement visual evidence and enhance temporal reasoning \cite{xiong2024large, fei2024vitron}.
As illustrated in Fig.~\ref{Fig:intension}, language provides additional cues about intent, ordering, and long-range dependencies, thereby offering discriminative semantics to resolve visually ambiguous cases.

Despite these advantages, however, current VLM-based TAL systems do not consistently outperform vision-only baselines \cite{huang2024vtimellm}. 
This paradox arises because VLMs frequently exhibit a \textit{modality bias}: instead of leveraging vision and language in a balanced manner, the inference is dominated by linguistic priors, causing visual evidence to be underutilized \cite{cadene2019rubi, schrodi2024two, chen2020counterfactual}. 

\begin{figure*}[t] 
    \centering
    \begin{adjustbox}{width=1\linewidth,center}
\includegraphics{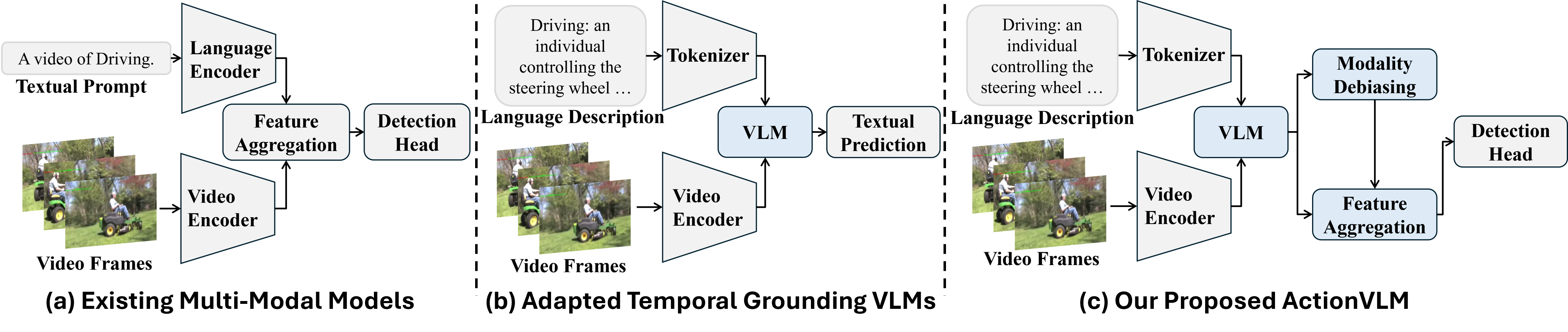} 
\end{adjustbox}
\caption{
(a) Existing multi-modal TAL models \cite{liberatori2024test,lin2022frozen} distribute uniform weights to vision and language.
(b) Language-centric VLMs predict in text.
(c) Our proposed ActionVLM mitigates modality bias by adaptively reweighting the language modality and preserving the vision fidelity during aggregation.
} 
\label{Fig:comparison} 
\end{figure*}

The emergence of this bias can be traced to several factors. 
(i) The \emph{attention sink} phenomenon, in which a subset of tokens or spatial locations consistently monopolizes attention mass regardless of task relevance \cite{xiao2023efficient}.
It is particularly harmful for fine-grained vision tasks, where informative signals are sparse and concentrated in a few vision tokens \cite{li2023redundancy}.
When those crucial tokens are suppressed, the model effectively becomes visually blind and overly reliant on language.
(ii) \emph{Language-centric} task formulations. As shown in Fig.~\ref{Fig:comparison}(b), VLMs are instructed and outputted in text formats, which encourages reliance on linguistic priors \cite{xiao2024can}.
When visual evidence is limited or dataset correlations induce strong linguistic priors, models tend to default to language, producing systematic errors and unwarranted overconfidence \cite{cadene2019rubi}.

Motivated by \citet{kang2025see} that redistribute attention weights to relieve sinks, we introduce a debiasing module that explicitly estimates the \emph{language advantage}—the incremental benefit of language over vision-only predictions—and adaptively reweights linguistic cues during fusion. 
Unlike prior debiasing approaches \cite{cadene2019rubi} that add extra branches and double computation, our design incurs negligible cost while ensuring language contributes only when it offers measurable gains, thus preventing attention sinks from suppressing critical visual evidence.
In parallel, inspired by studies highlighting the risks of language-centric task formulations \cite{xiao2024can}, 
we reformulate TAL with a residual feature aggregation strategy, where language is treated as a refinement signal rather than the primary driver. 
This vision-centric design maintains visual features as the backbone of temporal reasoning while leveraging language only as complementary semantics.  

Building on these two insights, we propose \textbf{ActionVLM}, a vision-centric framework for TAL that integrates explicit debiasing and residual aggregation to mitigate modality bias (Fig.~\ref{Fig:comparison}c). 
Our contributions are as follows:
\begin{itemize}[itemsep=0pt,topsep=0pt,parsep=0pt,leftmargin=10pt]
\item We identify modality bias as a fundamental challenge in applying VLMs to TAL and introduce a lightweight debiasing unit that adaptively reweights linguistic cues to counteract bias. 
\item Building on this, we propose a residual aggregation strategy that treats language as a complementary refinement signal, thereby enhancing temporal reasoning without compromising vision as the primary modality. 
\item We conduct extensive evaluations on three standard TAL benchmarks and demonstrate significant improvements over state-of-the-art methods.
\end{itemize}

\section{Related Work}
\label{sec:related_work}

\subsection{Temporal Action Localization}

Research on Temporal Action Localization (TAL) has traditionally been dominated by vision-only approaches.
Early methods relied on pre-extracted features from pretrained vision backbones \cite{carreira2017quo, chen2024internvl}, while recent approaches move toward end-to-end learning directly from raw videos \cite{liu2024end, yang2023basictad}, narrowing the gap between pretraining and downstream objectives. 
These features are typically fed into detection heads such as ActionFormer \cite{zhang2022actionformer}, which classify action presence at each time step and regress start-end boundaries.
Despite this progress, vision-only models remain constrained: actions with similar appearances are easily confused, resulting in inaccurate boundaries.

A central challenge is modeling temporal causality, namely the cause-effect dependencies across poses or events \cite{ning2019joint},
which is essential for action localization \cite{zhou2018temporal}. 
Yet it is hindered by vision redundancy, where informative cues are easily diluted by background \cite{li2023redundancy, wang2024retake}.
Consequently, models trained with only coarse visual supervision often overfit to spurious correlations. 
To address this, recent vision-only works attempt to strengthen temporal modeling.
ASL \cite{shao2023action} highlights discriminative frames via a Gaussian-based weighting scheme, while CausalTAD \cite{li2024causaltad} introduces causal modules. 
Nevertheless, their reliance on vision alone limits model's robustness.

Additionally, several TAL models \cite{liberatori2024test, lin2022frozen} have explored leveraging CLIP \cite{radford2021learning} to generalize to unseen actions, but this mainly benefits zero-shot capacity rather than localization performance. 
This is largely due to their fixed modality weights, which leave modality bias unresolved (Fig.~\ref{Fig:comparison}a).

\subsection{Language for Temporal Grounding}

Parallel works on temporal grounding demonstrate that language can provide complementary supervision \cite{lin2023univtg}. 
Language descriptions supply high-level semantics about action intent, ordering, and boundaries, offering explicit anchors that visual features often lack \cite{lei2019tvqa+}. 
Recent efforts adopt video-language models (VLMs). 
VTimeLLM \cite{huang2024vtimellm} aligns videos with captions via large-scale pretraining and adapts this alignment to instruction-based event localization. 
Other studies enhance temporal reasoning with explicit temporal indices \cite{ren2024timechat, chen2024timemarker} or embeddings \cite{qian2024momentor}. 

\subsection{Modality Bias}

Increasing evidence shows that VLMs often suffer from modality bias \cite{zheng2025mllms}, where models overfit to linguistic shortcuts and neglect visual cues.
This has been widely reported in vision-language tasks such as visual question answering, where removing visual features yields only marginal performance drops \cite{xiao2024can}. 

Prior work can be grouped into two directions. 
Diagnostic works primarily focus on revealing bias. 
For example, gradient-based analyses \cite{kwon2025see} and the attention sink studies \cite{xiao2023efficient} provide valuable insights into how linguistic shortcuts dominate inference but provide little in terms of mitigation.
In contrast, mitigation methods typically rely on either extra branches \cite{cadene2019rubi,clark2019don}, which introduce a language-only predictor to discount bias but double computation, or counterfactual data augmentations \cite{chen2020counterfactual}, which enforce visual reliance through handcrafted or synthetic mismatched pairs. 
Although effective, both are costly or unscalable, limiting large-scale video understanding.
\section{ActionVLM}

\begin{figure*}[t] 
    \centering
    \begin{adjustbox}{width=0.98\linewidth,center}
\includegraphics{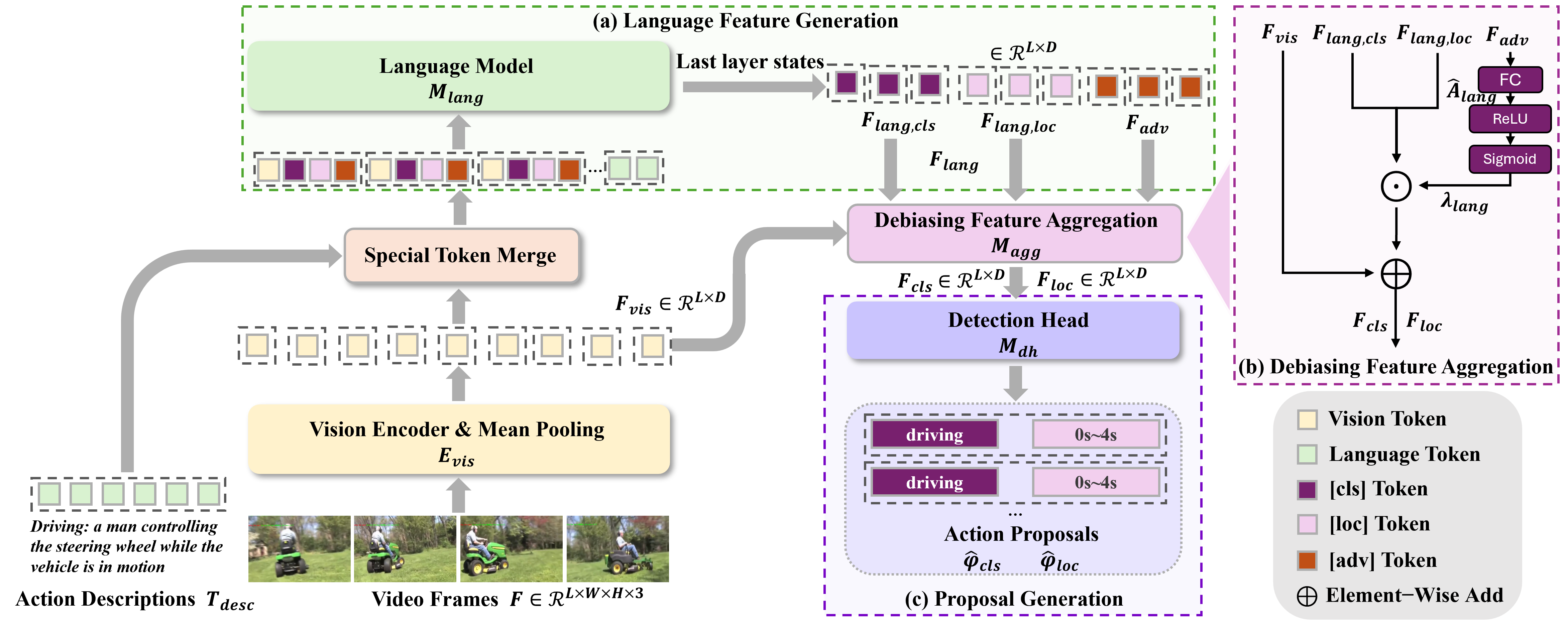} 
\end{adjustbox}
\caption{
The overview of our ActionVLM framework.
\textbf{(a) Language Feature Generation:} employ the language model to distill rich contextual information from action descriptions.
\textbf{(b) Debiasing Feature Aggregation:} scale the language features by estimating the contribution, then refine the vision features with reweighted language features.
\textbf{(c) Proposal Generation:} predict action proposals from the aggregated features.
} 
\label{Fig:framework} 
\end{figure*}

\subsection{Problem Definition and Motivation}

Given an untrimmed video with $L$ frames and $C$ predefined action categories, Temporal Action Localization (TAL) aims to produce frame-wise action proposals consisting of 
(1) a classification score sequence $\hat{\varphi}_{cls} \in \mathbb{R}^{L \times C}$, and
(2) a boundary localization sequence $\hat{\varphi}_{loc} \in \mathbb{R}^{L \times 2}$, where each tuple specifies the start and end timestamps.
A vision encoder $E_{vis}$ and a mean pooling layer map raw frames $F$ to per-frame vison features $F_{vis}=E_{vis}(F) \in \mathbb{R}^{L \times D}$, where $D$ is the feature dimension.

Recent works employ VLMs to inject language semantics, but still there is a gap between those models and strong vision-only baselines \cite{huang2024vtimellm}.
Formally, language models $M_{lang}$ are instructed and output in text, formulated as $(\hat{\varphi}_{cls},\hat{\varphi}_{loc}) = M_{lang}(F_{vis},T_{ins})$, where $T_{ins}$ denotes a task-specific instruction prompt, and the proposals $\hat{\varphi}_{cls}$ and $\hat{\varphi}_{loc}$ are embedded into a textual template.
This formulation shifts the learning burden onto the language component and encourages the system to explain visual inputs via linguistic shortcuts \cite{xiao2024can,cadene2019rubi}.
Consequently, visual signals can be down-weighted due to the attention sink \cite{xiao2023efficient,li2023redundancy}, which is especially detrimental to fine-grained temporal localization.

Instead of projecting actions into the linguistic embedding space, we treat the VLM as a semantic enhancer that injects language priors into visual representations.
Fig.~\ref{Fig:framework} illustrates the overview of ActionVLM.
To generate language features $F_{lang}$, we employ a generation module that distills semantics from action descriptions $T_{desc}$, defined as $F_{lang}=M_{lang}(F_{vis},T_{ins},T_{desc})$.
However, directly incorporating $F_{lang}$ risks amplifies modality bias toward language \cite{zheng2025mllms}. 
Thus, we introduce a debiasing mechanism, which estimates the language advantage $A_{lang}$, quantifying the additional information gained from language.
Only when $A_{lang}>0$, indicating complementary semantics rather than redundant bias, are the language features integrated into the visual stream. 
The fusion is performed by an aggregation module $M_{\mathrm{agg}}$, formulated as $(F_{cls},F_{loc})=M_{agg}(F_{vis},F_{lang},A_{lang})$, where the refined representations $F_{cls}$ and $F_{loc}$ are passed to a standard detection head $M_{\mathrm{dh}}$ \cite{zhang2022actionformer} to generate the action proposals: $(\hat{\varphi}_{cls},\hat{\varphi}_{loc}) = M_{dh}(F_{cls},F_{loc})$. 

\subsection{Language Feature Generation}
\label{sec:language_feature_generation}

Language feature generation leverages VLMs to inject semantic knowledge from action descriptions $T_{desc}$.
Unlike weak visual supervision without annotating the temporal causality, action descriptions can explicitly encode cause-effect relations across ordered poses.
As shown in Fig.~\ref{Fig:intension}, those linguistic supervisions provide higher-level guidance that helps disambiguate visually similar actions.
In practice, such descriptions can be easily obtained either from human experts or intelligent models. 
We adopt GPT-4o \cite{gpt4o} to generate action descriptions, prompting it to specify representative start and end poses for each action without access to visual inputs, detailed in Appx.~\ref{appx:prompt_descriptions}.

Since VLMs process in discrete tokens while downstream tasks require continuous representations, we follow \citet{yuan2025sa2va} and extract the last-layer hidden states of special tokens to construct continuous features.
Specifically, three special tokens are introduced, namely [cls], [loc] and [adv].  
[cls] and [loc] summarize information for action classification and localization, respectively, while [adv] estimates the contribution of the language modality. 
As illustrated in Fig.~\ref{Fig:framework}(a), each video frame is represented by its visual feature $F_{vis}^{(l)}$ and three special tokens, with the action description tokens $T_{desc}$ appended to the end of the sequence.

After VLM processing, contextual language priors and critical visual cues are propagated throughout the sequence. 
The special tokens are trained to summarize target-specific information under this enriched context. 
Their final hidden states are extracted as continuous features, namely $F_{lang,cls}, F_{lang,loc},F_{adv} \in \mathbb{R}^{L \times D}$, which serve as adaptable inputs for downstream modules.

\subsection{Language Advantage}

Language can benefit TAL by providing high-level semantics difficult to infer from visual evidence alone. 
However, because TAL is fundamentally a vision-centric task, language can also become a shortcut: the model may reduce the training objective by exploiting frequent textual patterns rather than grounding its prediction in the video content. 
As a result, the multimodal loss alone is not a reliable signal of whether language is truly helpful.

To make this distinction explicit, we compare the task loss of a vision-only branch, denoted as $\mathcal{L}_{v}$, with that of the corresponding vision-language branch, denoted as $\mathcal{L}_{vl}$. 
Here, $\mathcal{L}_{v}$ reflects how well the task can be solved from visual evidence alone, while $\mathcal{L}_{vl}$ measures the loss after language is incorporated. 
Based on this comparison, we define Language Advantage (LA) as $A_{lang}=\mathcal{L}_{v}-\mathcal{L}_{vl}$.

This quantity measures the utility of language relative to vision. 
When language provides complementary semantics and improves prediction beyond the visual baseline, we have $\mathcal{L}_{vl} < \mathcal{L}_{v}$ and thus $A_{lang} > 0$. 
When language introduces misleading priors or overrides visual evidence, $\mathcal{L}_{vl}$ can even exceed $\mathcal{L}_{v}$, yielding a small or negative $A_{lang}$ and indicating harmful modality bias.

This definition also explains why directly optimizing the multimodal branch can be problematic. 
In conventional VLMs, the model is trained only through $\mathcal{L}_{vl}$, so any shortcut-induced loss reduction would create a positive feedback loop that further increases reliance on language. 
In contrast, LA uses the vision-only branch as a stable reference, allowing us to determine whether language is providing genuine complementary information or merely acting as a shortcut. 

\textbf{Alternating-Epoch Estimation.}
A naive implementation would require an additional vision-only branch to compute $\mathcal{L}_{v}$ at every iteration. To avoid this overhead, we adopt a lightweight alternating-epoch estimation inspired by reinforcement learning \cite{schulman2015high}. Training alternates between vision-only and vision-language epochs. Vision-only epochs compute class-wise average losses $\bar{\mathcal{L}}_v^{(c)}$, which then serve as a stable proxy for $\mathcal{L}_v$ during vision-language epochs. The language advantage for the $l^{\text{th}}$ frame is thus estimated as
\begin{equation}
\label{eq:adv_lv_estimation}
A_{lang}^{(l)}=\bar{\mathcal{L}}_{v}^{(c)}-\mathcal{L}_{vl}^{(l)}, l \in \{1,2,...,L\},
\end{equation}
where $L$ is the number of frames, and $c$ the ground-truth action category of that frame. 

\begin{algorithm}[t]
\small
\caption{Training Pipeline}
\label{algo:training}
\KwIn{Dataset $\mathcal{D}$; number of frames $L$;number of classes $C$;  number of samples in class $c$: $N^{(c)}$; ground truth class $\varphi_{cls}$}
\noindent\rule{0.93\linewidth}{0.7pt} 
\textbf{\# Vision-Only Epoch}\\
Initialize class-wise loss $\bar{\mathcal{L}}_v^{(c)} \leftarrow 0$ for all actions\;
\ForEach{$(F_{vis}, \varphi_{cls}, \varphi_{loc}) \sim \mathcal{D}$}{
    Compute task loss $\mathcal{L}_v$ \textbf{on vision features} $F_{vis}$\;
    Accumulate mean loss per class:
    $\bar{\mathcal{L}}_v^{(\varphi_{cls})} \! += \! \frac{\mathcal{L}_v}{N^{(\varphi_{cls})}}$\;
Backpropagate vision-only task loss: $\mathcal{L}_v$\;
}
\noindent\rule{0.93\linewidth}{0.7pt} 
\textbf{\# Vision-Language Epoch}\\
    \ForEach{$(F_{vis}, F_{lang}, F_{adv}, \varphi_{cls}, \varphi_{loc}) \sim \mathcal{D}$}{
        Estimate advantage by Eq.~\ref{eq:adv_estimation}:
        $\hat{A}_{lang} = FC(F_{adv})$\;
        Aggregate $F_{vis}$ and $F_{lang}$ by Eq.~\ref{eq:lamda_l} and \ref{eq:feat_agg} to obtain vision-language features $F_{cls}$ and $F_{loc}$\;
        Compute task loss $\mathcal{L}_{vl}$ and text generation loss $\mathcal{L}_{tg}$ \textbf{on vision-language features} $F_{cls}$ and $F_{loc}$ by Eq.~\ref{eq:loss_detection} and \ref{eq:loss_text}\;
        Estimate target advantage by Eq.~\ref{eq:adv_lv_estimation}: 
        $A_{lang}^{(l)}=\bar{\mathcal{L}}_{v}^{(\varphi_{loc})}-\mathcal{L}_{vl}^{(l)}$\;
        Compute advantage regression by Eq.~\ref{eq:adv_loss}:
        $\mathcal{L}_{adv}=\frac{1}{L}\sum^L_{l=1}(\hat{A}_{lang}^{(l)}-A_{lang}^{(l)})^2$\;
        Backpropagate total loss by Eq.~\ref{eq:all_loss}:
    $\mathcal{L} = \mathcal{L}_{vl} + \lambda_{adv}\mathcal{L}_{adv} + \lambda_{tg}\mathcal{L}_{tg}$\;

    }
\end{algorithm}

To predict LA at inference, the learnable [adv] tokens are trained to regress the approximated advantage from their hidden states $F_{adv}$ by
\begin{equation}
\label{eq:adv_estimation}
\hat{A}_{lang}=FC(F_{adv}),
\end{equation}
where FC denotes a fully connected projection.
We then define the advantage loss as
\begin{equation}
\label{eq:adv_loss}
\mathcal{L}_{adv}=\frac{1}{L}\sum^L_{l=1}(\hat{A}_{lang}^{(l)}-A_{lang}^{(l)})^2
\end{equation}

Algorithm~\ref{algo:training} summarizes our alternating-epoch estimation of LA. 
During each vision-only epoch, we compute the task loss using only visual features and accumulate the class-wise average loss $\bar{\mathcal{L}}_v^{(c)}$. 
In the subsequent vision-language epoch, this quantity serves as a lightweight proxy for the vision-only loss $\mathcal{L}_v$.
In this way, LA is anchored to a vision-only baseline without requiring an additional vision branch at every iteration.

This estimation is computationally efficient for three reasons. First, the two training phases share the same backbone and differ only in whether the language model is activated, avoiding architectural duplication. Second, vision-only epochs skip the language model entirely and are therefore cheaper than standard multimodal updates, while still providing a principled reference for calibrating modality contributions. Third, because LA is derived directly from task losses rather than dataset-level bias statistics, it is naturally model- and dataset-agnostic, making the same estimation strategy applicable across different TAL benchmarks and VLM backbones.

\subsection{Debiasing Feature Aggregation}

To dynamically regulate the contribution of language, we derive a reliability weight $\lambda_{lang}$ from the predicted LA $\hat{A}_{lang}$, using it as a utility signal to determine whether language provides information beyond vision and should therefore participate in vision-grounded localization.
To avoid assigning positive weights to negative advantages, we clip it and scale it into $[0,1)$ through sigmoid mapping:
\begin{equation}
\label{eq:lamda_l}
\lambda_{lang}=2\sigma(ReLU(\hat{A}_{lang}))-1
\end{equation}

Rather than relying solely on language features for downstream tasks, we preserve the visual feature $F_{vis}$ and treat language features as a bounded residual refinement. Inspired by residual learning~\cite{he2016deep}, this design keeps vision dominant and allows language to complement it only when beneficial, thereby mitigating modality bias.
As illustrated in Fig.~\ref{Fig:framework}(b), the final representations for classification and localization are obtained via element-wise addition:
\begin{equation}
\label{eq:feat_agg}
\begin{aligned}
F_{cls} &= F_{vis} \oplus \lambda_{lang}F_{lang,cls} \\
F_{loc} &= F_{vis} \oplus \lambda_{lang}F_{lang,loc}
\end{aligned}
\end{equation}

\begin{table*}[b]
\centering
\begin{adjustbox}{width=1\textwidth,center}
\begin{threeparttable}
\begin{tabular}{lll|cccc|cccccc}
\hline
                                         &  & &  \multicolumn{4}{c|}{\textbf{ActivityNet-1.3}}  &  \multicolumn{6}{c}{\textbf{THUMOS14}} \\
\textbf{Model}        & \textbf{Vision} & \textbf{Language} & \textbf{0.5}  & \textbf{0.75}   &  \textbf{0.95} & \textbf{Avg.}  & \textbf{0.3}  & \textbf{0.4}  & \textbf{0.5} & \textbf{0.6}  & \textbf{0.7} & \textbf{Avg.}\\
\hline
\textit{\textbf{pre-extracted}}&&&&&&&&&&&&\\

ASL  \cite{shao2023action}      &TSP/I3D       &  -        & 54.1           & 37.4           & 8.0            & 36.2           & 85.3          & 80.9          & 73.4         & 61.1          & 45.1         & 69.2 \\
Pred-DETR \cite{kim2025prediction}& I3D/I3D    &  -        & 58.4           & 39.1           & \underline{9.9}    & 38.6           & 84.1          & 80.0          & 72.2         & 60.4          & 45.8         & 68.5 \\
DiGIT \cite{kim2025digit}       & R(2+1)D/I3D  &  -        & 54.4           & 38.2           & \textbf{10.7}  & 37.3           & 83.6          & 79.6          & 71.9         & 61.5          & 48.6         & 69.0 \\
ADI-Diff \cite{foo2024action}   &R(2+1)D/I3D   &  -        & 56.9           & 38.9           & 9.1            & 38.3           & 84.9          & 81.5          & 76.5         & 63.0          & 48.0         & 70.8 \\
ActionFormer \cite{zhang2022actionformer}&TSP/VM2-g&  -       & 55.1           & 38.3        & 8.9            & 37.1        & 84.0          & -             & 73.0         & -             & 47.7         & 69.6 \\
MFAM \cite{tang2024learnable}   &  I3D/VM2-g     &  -        & 54.7           & 37.3         & 8.6            & 36.6         & 84.6          & 80.8          & 73.5         & 61.7          & 48.6         & 69.8 \\
TriDet  \cite{shi2023tridet}    &TSP/VM2-g       &  -        & 54.7           & 38.0         & 8.4            & 36.8         & 84.8          & -             & 73.3         & -             & 48.8         & 70.1 \\
DyFADet \cite{yang2024dyfadet}  & TSP/VM2-g      &  -        & 58.1           & 39.6         & 8.4            & 38.5         & 86.0          & 81.7          & 76.3         & 64.5   & 50.1         & 71.7 \\
\textbf{ActionVLM-1.5B (ours)}& TSP/VM2-g      & QW2.5-1.5B& \underline{58.6}   & \underline{39.7}   & 9.3            & \underline{39.0} & \underline{87.3}  & \underline{82.6}           & \underline{76.9}            & \underline{64.9}           & \underline{50.7} & \underline{72.5} \\
\textbf{ActionVLM-3B (ours)}  & TSP/VM2-g      &  QW2.5-3B & \textbf{58.8}  & \textbf{40.0}& 9.5  &\textbf{39.3} &\textbf{87.7}  &\textbf{83.4}  & \textbf{77.4}   & \textbf{65.2} & \textbf{50.9} & \textbf{72.9} \\
\hline
\textit{\textbf{end2end - small vision encoder}}&&&&&&&&&&&&\\
ViT-TAD \cite{yang2024adapting} &  VM2-S            &  -        & 55.1           & 37.8           & 8.8            & 36.7           & 79.8          & 75.2          & 69.4         & 56.4          & 41.7         & 64.3\\
Re$^{2}$TAL \cite{zhao2023re2tal}&  Re$^{2}$Swin-T  &  -        & 55.1           & 37.1           & 8.3            & 36.5           & 82.1          & 78.1          & 60.1         & 59.8          & 44.4         & 66.9\\
AdaTAD  \cite{liu2024end}       & VM-S      &  -        & 56.2           & 39.0           & 9.1            & 37.8           & 84.5          & 80.2          & 71.6         & 60.9          & 46.9         & 68.8    \\
\textbf{ActionVLM-S-1.5B (ours)}&VM2-S&QW2.5-1.5B     &\underline{57.8}        & \underline{39.9}           & \underline{9.4}            & \underline{38.9}       &\underline{86.6}      &\underline{82.3}      &\underline{75.0}      & \underline{63.9}          & \underline{49.0}         & \underline{71.4} \\
\textbf{ActionVLM-S-3B (ours)}&VM2-S&QW2.5-3B &\textbf{58.1}&\textbf{40.2}&\textbf{9.6}&\textbf{39.2}&\textbf{87.2}&\textbf{82.9}&\textbf{75.8}&{\textbf{64.6}}&\textbf{49.6}&\textbf{72.0}\\
\hline
\textit{\textbf{end2end - large vision encoder}}&&&&&&&&&&&&\\
ViT-TAD \cite{yang2024adapting}   & VM2-B          &  -    & 55.9           & 38.5           & 8.8            & 37.4           & 85.1          & 80.9          & 74.2         & 61.8          & 45.4         & 69.5\\
Re$^{2}$TAL \cite{zhao2023re2tal}&  Re$^{2}$SF-101 &  -    & 55.8           & 38.5           & 9.4            & 37.6           & 84.2          & 80.3          & 74.3         & 62.7          & 49.6         & 70.2\\
AdaTAD   \cite{liu2024end}        & VM-S           &  -    & 56.8           & 39.4           & 9.7            & 38.4           & 87.0          & 82.4          & 75.3         & 63.8          & 49.2         & 71.5     \\
\textbf{ActionVLM-B-1.5B (ours)}&VM2-B&QW2.5-1.5B&\underline{58.4}    &\underline{40.3}  &\underline{9.8}       & \underline{39.4}           &\underline{88.1}       &\underline{84.5}       &\underline{77.4}      & \underline{66.9}         & \underline{51.5}         &\underline{73.7}\\
\textbf{ActionVLM-B-3B (ours)}& VM2-B&QW2.5-3B&\textbf{58.8}&\textbf{40.6}  &\textbf{10.2} &\textbf{39.7}&\textbf{88.8}&\textbf{85.1}&\textbf{78.2}&\textbf{67.1} &\textbf{51.8}&\textbf{74.2}\\

\hline
\end{tabular}
\end{threeparttable}
\end{adjustbox}
\caption{Results of mAPs (\%) at different IoU thresholds on THUMOS14 and ActivityNet-1.3.
The first and second best results for each pipeline are highlighted in \textbf{bold} and \underline{underlined}.}
\label{table:benchmark}
\end{table*}

\subsection{Detection Head}
\label{sec:rpn}

We adopt the Detection Head (DH) of ActionFormer~\cite{zhang2022actionformer}, which employs 1D convolutions to transform the aggregated features (Eq.~\ref{eq:feat_agg}) into action proposals $\hat{\varphi}_{cls}$ and $\hat{\varphi}_{loc}$.
The detection head loss is defined as
\begin{equation}
\label{eq:loss_detection}
\mathcal{L}_{dh} = \frac{1}{M}\sum_{l=1}^L 
\left(
\mathcal{L}_{cls} + 
\lambda_{loc}\mathbf{1}_{\mathcal{C}_l}\mathcal{L}_{loc}
\right),
\end{equation}
where $\mathcal{L}_{cls}$ is the focal loss~\cite{tian2019fcos} for classification, $\mathcal{L}_{loc}$ is the DIoU loss~\cite{rezatofighi2019generalized} for localization, $M$ is the number of proposals, $L$ is the frame length, $\mathbf{1}_{\mathcal{C}_l}$ indicates positive frames, and $\lambda_{loc}$ balances localization.

\subsection{Training Objectives}

To further align the language model with the TAL objective, we retain textual action proposal generation as an auxiliary task.
It generates a sentence $\hat{T}_{pred}$ listing the ground truth proposals:
\begin{equation}
\label{eq:vlm_text}
\hat{T}_{pred} = M_{lang}(F_{vis}, T_{ins},T_{desc})
\end{equation}

We optimize it using a cross-entropy loss:
\begin{equation}
\label{eq:loss_text}
\mathcal{L}_{tg}=L_{ce}(\hat{T}_{pred},(\varphi_{cls},\varphi_{loc})),
\end{equation}
where the ground truths $(\varphi_{cls},\varphi_{loc})$ are formatted into a structured textual template.


The overall training objective combines DH loss $\mathcal{L}_{dh}$, the text generation loss $\mathcal{L}_{tg}$ (Eq.~\ref{eq:loss_text}), and the  advantage loss $\mathcal{L}_{adv}$ (Eq.~\ref{eq:adv_loss}):
\begin{equation}
\label{eq:all_loss}
\mathcal{L}=\mathcal{L}_{dh}+\lambda_{tg}\mathcal{L}_{tg}+\lambda_{adv}\mathcal{L}_{adv},
\end{equation}
where $\lambda_{tg}$ and $\lambda_{adv}$ control the relative weights.

\section{Experiment}
\label{sec:exp}

\subsection{Experimental Setting}
\textbf{Datasets.} We evaluate our method on THUMOS14 \cite{idrees2017THUMOS} and ActivityNet-1.3 \cite{caba2015activitynet}.
THUMOS14 and ActivityNet are third-person benchmarks, consisting of 413 and 19,994 videos, respectively.
In appendix, we also evaluate on EPIC-Kitchens 100 \cite{damen2020epic}, which consists of 700 egocentric videos. 

\noindent
\textbf{Metrics.}  We report mean Average Precision (mAP) at various temporal Intersection over Union (tIoU) thresholds, following the common practice \cite{liu2019completeness}.
tIoU is defined as the temporal overlap.
At each tIoU threshold, mAP is computed by averaging the precision across actions.

\noindent
\textbf{Training.}
Our training involves three components. 
The detection head is trained from scratch with a learning rate (lr) of 1e-4.
The vision encoder we use is the pre-trained VideoMAEv2 (VM2) \cite{wang2023videomae}, which is further fine-tuned in our framework with an lr of 5e-6. 
The language model is fine-tuned using the LoRA \cite{hu2022lora} with a rank of 128, an alpha of 256, and an lr of 5e-6.
All training uses DeepSpeed \cite{rasley2020deepspeed} with AdamW on an Nvidia L40 for 120 epochs. 
The batch size is set to 1 for THUMOS14 and 4 for ActivityNet.
Following \citet{zhang2022actionformer}, we set the $\lambda_{loc}$ (Eq.~\ref{eq:loss_detection}) to 1.
The auxiliary task weights $\lambda_{tg}$ and $\lambda_{adv}$ in Eq.~\ref{eq:all_loss} are empirically set to 0.1.

\subsection{Comparison with SoTA Methods}
Table~\ref{table:benchmark} summarizes results on two standard TAL benchmarks.
ActionVLM consistently attains state-of-the-art performance across all training regimes.
Among end-to-end models, ActionVLM-B-3B reaches 74.2\% mAP on THUMOS14 and 39.7\% on ActivityNet-1.3. 
Compared with strong vision-only baselines built on the same vision backbone, ActionVLM-S surpasses ViT-TAD by 7.7\% mAP on THUMOS14 and 2.5\% on ActivityNet.
Even with frozen pre-extracted features, our method yields a 1.2\% gain over DyFADet. 

These aggregate numbers indicate that our ActionVLM possesses universality that transcends specific architectures or datasets.
The behavior aligns with our analysis in Appx.~\ref{appx:visualization_ambiguity}, where ActionVLM provides clearer benefits when visual evidence is ambiguous, a common challenge in TAL. 
This suggests that language contributes primarily by reinforcing semantic discrimination, providing stronger semantic cues when uncertain.

To examine robustness to language model capacity, we evaluate two InternVL2.5 \cite{chen2024internvl} variants that use Qwen2.5-1.5B and 3B as their language backbones. 
InternVL3 is chosen for its strong alignment with visual downstream tasks.
ActionVLM-S-3B achieves a 0.6\% mAP improvement over the 1.5B counterpart on THUMOS14, suggesting that broader knowledge and richer pretraining better support TAL.


\begin{table*}[t]
\centering
\begin{adjustbox}{width=1\textwidth,center}
\begin{tabular}{cl|cccccc|ccc}
\hline
 & &  \multicolumn{9}{c}{\textbf{THUMOS14}}                                                                    \\ 
\textbf{Task}           & \textbf{Model}                                                       &\textbf{0.3} & \textbf{0.4}& \textbf{0.5}& \textbf{0.6}& \textbf{0.7}&\textbf{Avg.}&\textbf{Fixed$\downarrow$}&\textbf{Infinite$\downarrow$}&\textbf{LAP$\downarrow$}\\
\hline
\multirow{2}{*}{QA}                 &VTimeLLM \cite{huang2024vtimellm}        &57.4         &46.7         &33.8         &19.7         & 9.6         & 33.4        & 14.6 &  27.7  &    14.8  \\
                                    &Timemarker \cite{chen2024timemarker}     &63.8         &55.0         &43.6         &33.0         & 18.9        & 42.9     & 8.7 &  11.2  &   16.5   \\
\hline
\multirow{3}{*}{FA}&ActionVLM (only VLM features $F_{lang}$ w/o $F_{vis}$ in Eq.~\ref{eq:feat_agg})&75.1 & 70.7 & 61.4  &54.9         & 34.4        & 59.3      &  0 &   0  &  8.8    \\
                                    &ActionVLM (only vision features $F_{vis}$ w/o $F_{lang}$ in Eq.~\ref{eq:feat_agg})& 85.0        & 80.4        & 73.5        & 63.1        & 47.8        & 69.9 & 0  & 0  &   -  \\
                                    &\textbf{ActionVLM}(full, with the aggregation of $F_{lang}$ and $F_{vis}$) &\textbf{87.2}&\textbf{82.9}&\textbf{75.8}&\textbf{64.6}&\textbf{49.6}&\textbf{72.0}& 0  & 0  & \textbf{2.6} \\
\hline
\end{tabular}
\end{adjustbox}
\caption{Quantifying modality bias using Language Advantage on Performance (LAP) and two hallucination rates (\%) under two task formulations: Question-Answering (QA) and our Feature Aggregation (FA) approach.}
\label{table:modality_bias_task_formulation}
\end{table*}

\begin{figure*}[t] 
    \centering
    \begin{adjustbox}{width=0.96\linewidth,center}
\includegraphics{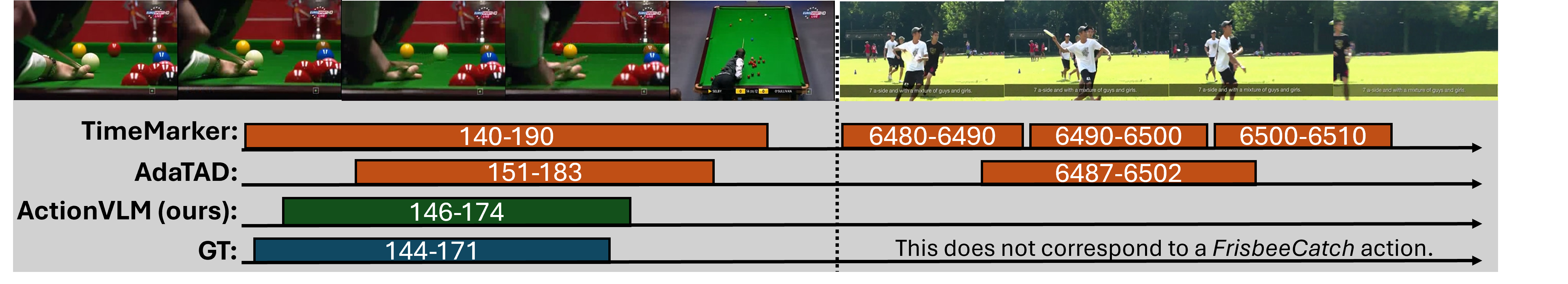} 
\end{adjustbox}
\caption{
Visualization of modality bias in localizing two challenging actions, \textit{Billiard (left)} and \textit{FrisbeeCatch (right)}. 
We compare our ActionVLM with a question-answering-based model (TimeMarker) and a vision-only model (AdaTAD). 
Numbers inside the bars denote the corresponding frame indices.
} 
\label{Fig:modality_bias} 
\end{figure*}

\begin{table*}[t]
\centering
\begin{adjustbox}{width=0.81\textwidth,center}
\begin{tabular}{l|ccc|ccc|ccc}
\hline
& \multicolumn{3}{c|}{\textbf{Easy}}    & \multicolumn{3}{c|}{\textbf{Medium}}  & \multicolumn{3}{c}{\textbf{Hard}}                                                                  \\ 
\textbf{Model}                                             &\textbf{mAP}&\textbf{Gain}&\textbf{mLA}&\textbf{mAP}&\textbf{Gain}&\textbf{mLA}&\textbf{mAP}&\textbf{Gain}&\textbf{mLA}\\
\hline
AdaTAD (baseline)                                             & \textbf{93.2}& -         & -          &76.1        & -           & -          &31.1        & -             & - \\
Timemarker                                                    & 71.0       & -22.2       & -          &47.2        & -28.9       & -          &13.7        & -17.4         & - \\     
\hline
ActionVLM (learnable $\lambda_{lang}$ w/o $\mathcal{L}_{adv}$)& 86.6       & -6.6        & 0.16       &78.5       & 2.4          & 0.20       &35.4        & 4.3          & 0.23 \\
\textbf{ActionVLM}(full, with $\mathcal{L}_{adv}$)            &89.3&-3.9&0.08 &\textbf{78.6}&\textbf{2.5}&0.19 &\textbf{40.1}&\textbf{9.0}&0.36  \\
\hline
\end{tabular}
\end{adjustbox}
\caption{Performance breakdown across different visual ambiguity level (mLA denotes the average frame-level LA).}
\label{table:modality_bias_difficulty}
\end{table*}

\begin{table*}[htbp]
\centering
\begin{adjustbox}{width=0.83\textwidth,center}\begin{tabular}{l|cccccc|cc}
\hline
 &  \multicolumn{8}{c}{\textbf{THUMOS14}}                                                                    \\ 
\textbf{Model}                               &\textbf{0.3} &\textbf{0.4} &\textbf{0.5} &\textbf{0.6} &\textbf{0.7} &\textbf{Avg.} &\textbf{\textit{Billiard}} &\textbf{\textit{FrisbeeCatch}}\\
\hline
ActionVLM (fixed $\lambda_{lang}=1.0$ )      & 84.5        & 79.9        & 72.7        & 62.0        & 46.7        & 69.2 & 34.8   & 26.7 \\
ActionVLM (fixed $\lambda_{lang}=0.8$ )      & 85.0        & 80.4        & 73.1        & 62.5        & 46.8        & 69.6 & 35.9   & 27.7 \\
ActionVLM (fixed $\lambda_{lang}=0.6$ )      & 85.6        & 81.1        & 73.6        & 63.2        & 47.5        & 70.2 & 38.4   & 30.9 \\
ActionVLM (fixed $\lambda_{lang}=0.4$ )      & 86.6        & 81.8        & 74.2        & 63.6        & 48.0        & 70.8 & 39.1   & 31.9 \\
ActionVLM (fixed $\lambda_{lang}=0.2$ )      & 86.2        & 81.6        & 74.0        & 63.5        & 47.9        & 70.6 & 38.2   & 31.7 \\
ActionVLM (fixed $\lambda_{lang}=0.0$ )      & 85.0        & 80.4        & 73.5        & 63.1        & 47.8        & 69.9 & 35.3   & 27.4 \\
\hline
ActionVLM (LLM-generated $\lambda_{lang}$ w/o $\mathcal{L}_{adv}$)& 85.8 & 81.5  & 73.3  &  63.0     & 47.1        & 70.1 & 32.0   & 32.0 \\
ActionVLM (learnable $\lambda_{lang}$ w/o $\mathcal{L}_{adv}$)& 86.1     & 81.6  & 74.1  &  63.7     & 48.2        & 70.7 & 38.2   & 32.5 \\
\textbf{ActionVLM}(full, with $\mathcal{L}_{adv}$)&\textbf{87.2}&\textbf{82.9}&\textbf{75.8}&\textbf{64.6}&\textbf{49.6}&\textbf{72.0}&\textbf{44.5}&\textbf{35.7}\\
\hline
\end{tabular}
\end{adjustbox}
\caption{Comparison of mAPs and the AP of two visually ambiguous actions under ablations of different language modality weights $\lambda_{\text{lang}}$ (Eq.~\ref{eq:feat_agg}) and the language advantage loss $\mathcal{L}_{\text{adv}}$ (Eq.~\ref{eq:adv_loss}).}
\label{table:lang_weight}
\end{table*}

\subsection{Analysis on Modality Bias}
\label{sec:exp:modality_bias}
\textbf{Measuring Modality Bias.}
To quantify modality bias at the performance level, we adopt Language Advantage on Performance (LAP), inspired by the diagnostic protocol of \citet{cadene2019rubi}. 
Unlike the loss-based LA used during training, LAP evaluates how much prediction quality deteriorates when language priors conflict with the visual content. 
Specifically, we replace the aligned action descriptions with mismatched ones and measure the relative drop in average mAP. 
To create conflicts, we preferentially swap descriptions between semantically similar actions. 
When no clear counterpart exists, we randomly shuffle the description-action pairs. 
A larger LAP indicates that the model is more easily misled by text and therefore more biased toward linguistic cues, whereas a smaller LAP indicates stronger reliance on visual evidence and better robustness to misleading language.

We further examine two extreme hallucination patterns caused by modality bias.
1) Fixed outputs refer to nearly identical temporal segments repeatedly predicted within a sliding video window regardless of the input content, which we detect by checking whether segment intervals remain unchanged across consecutive steps. 
2) Infinite listing refers to the model repeatedly emitting previously predicted segments without producing an End-of-Sequence token, which we detect by matching tail segments against earlier predictions. 
As shown in Table~\ref{table:modality_bias_task_formulation}, QA-based VLMs exhibit both high LAP and non-trivial hallucination rates, whereas our FA formulation reduces LAP to 2.6 and eliminates both failure modes, indicating substantially stronger visual grounding.

\textbf{Question-Answering (QA) Formulation.}
As multimodal TAL has rarely been studied beyond zero-shot settings \cite{liberatori2024test, lin2022frozen}, systematic analysis of modality bias in TAL is lacking. 
To this end, we adapt two recent temporal grounding VLMs—VTimeLLM \cite{huang2024vtimellm} and Timemarker \cite{chen2024timemarker}—to the TAL setting using a QA formulation.

Our analysis reveals a fundamental limitation of this paradigm. 
As illustrated in Fig.~\ref{Fig:modality_bias}, Timemarker often produces implausible but linguistically preferred boundaries on challenging videos (e.g., coarse rounded timestamps or obviously invalid segments), revealing a reliance on common numerical patterns rather than frame-level evidence.

Quantitatively, Table~\ref{table:modality_bias_task_formulation} shows that QA-based VLMs exhibit substantial degradation with up to 16.5\% LAP, suggesting a strong modality bias.
Such sensitivity confirms that QA formulations, which generate ranked textual proposals, inherently overemphasize linguistic patterns and weaken visual grounding, making them ill-suited for TAL.

\textbf{Feature Aggregation (FA) Formulation.}
Instead of casting TAL as a text-generation problem, our Feature Aggregation (FA) formulation avoids LLM explicitly predicting boundaries in language space, and fundamentally removes the structural source of hallucinations observed in QA-based models (Table~\ref{table:modality_bias_task_formulation}).

However, Table~\ref{table:modality_bias_task_formulation} reveals that directly using VLM-derived language features $F_{lang}$ still underperforms the vision-only variant by 10.6\% mAP. 
This gap highlights that features from language-centric VLMs remain insufficient for dense temporal grounding due to the inherent modality bias.

Our full ActionVLM resolves this limitation through our debiasing FA.
By explicitly restoring vision as the dominant signal and using language only as a complementary prior, ActionVLM achieves the best overall performance while reducing LAP to 2.6\%. 
As illustrated in Fig.~\ref{Fig:modality_bias}, language priors attend to semantically relevant cues (e.g., subtle arm motions in \textit{Billiards}), enabling more precise localization than vision-only AdaTAD, which relies primarily on global visual saliency. 
This demonstrates that FA leverages language to refine, rather than override, visual grounding.

\subsection{Analysis on Language Advantage}
\label{sec:exp:language_advantage}
To analyze how Language Advantage (LA) affects performance, we group action categories in THUMOS14 into easy, medium, and hard subsets based on the performance of the vision-only baseline (detailed in Appx.~\ref{appx:action_category}).
The hard subset includes \textit{Billiard} and \textit{FrisbeeCatch}. 
While multiple factors may contribute to their lower performance, both actions exhibit notable visual ambiguity:
\textit{Billiard} exhibits subtle and sparse motion cues, while \textit{FrisbeeCatch} is easily confused with throwing actions and differs only in temporal pose order (Fig.~\ref{Fig:modality_bias}).

Table~\ref{table:modality_bias_difficulty} reveals three key observations.
First, higher visual ambiguity is associated with stronger modality bias. 
When visual cues are limited, models lack sufficient visual evidence and consequently tend to linguistic priors from pretraining, leading to pronounced degradation on vision tasks.

Second, ActionVLM alleviates modality bias via LA-driven adaptive weighting.
For hard cases, ActionVLM assigns higher mLA, allowing language to provide complementary semantics when vision struggles.
The strong positive correlation between LA and performance gains in Table~\ref{table:modality_bias_difficulty} confirms that LA reliably identifies bias-sensitive cases and activates language selectively rather than uniformly.

Third, the advantage loss $\mathcal{L}_{adv}$ (Eq.~\ref{eq:adv_loss}) is essential for suppressing redundant language cues.
Without $\mathcal{L}_{adv}$, the model assigns non-trivial language weights to easy actions (mLA=0.16), where vision is already sufficient, resulting in performance drops (-6.6 mAP vs. AdaTAD).
The full model learns pseudo-optimal modality weights, reducing mLA to 0.08 on easy actions while amplifying language contributions for ambiguous ones, which explains why performance gains concentrate on hard cases.

Table~\ref{table:lang_weight} further compares ActionVLM under different settings of the language weight. 
With fixed or globally learnable weights, models yield only marginal improvements and can even degrade performance when language is overemphasized—confirming that language provides complementary cues in vision tasks while excessive reliance introduces bias. 
This also highlights that language is beneficial only when adaptively invoked, and effective modality debiasing in TAL requires sample-specific weighting rather than static fusion.

\section{Conclusion}

In this study, we present ActionVLM, a vision-centric VLM for Temporal Action Localization (TAL) that explicitly addresses modality bias. 
Rather than allowing language to dominate prediction through shortcut correlations, our framework preserves visual grounding and invokes language only when it provides measurable benefit, through language advantage estimation and debiased residual aggregation. 
In this way, ActionVLM enables language to refine rather than override visual grounding.
Extensive experiments show that this design not only alleviates modality bias and improves robustness under visual ambiguity, but also yields consistent gains across TAL benchmarks with negligible overhead. 
More broadly, our results suggest a practical multimodal alternative to purely visual scaling, pointing toward robust video understanding by better calibrating language priors rather than relying solely on ever larger visual data and models. 


\section*{Limitations}
While ActionVLM demonstrates strong performance, it has several limitations. 
First, incorporating vision-language models increases memory usage. Our small-sized variant, however, can be trained end-to-end on 24 GB GPUs, which covers most widely available hardware. Future work may explore lighter adaptations to further reduce resource costs.
Second, following prior works \cite{zhang2022actionformer}, we adopt a sliding-window dataloader with a fixed window size, which may introduce label noise by treating partial clips as complete actions. More adaptive loading strategies could improve robustness.
Third, language descriptions may introduce inherent bias due to imperfect alignment with fine-grained visual cues. 
However, our debiasing feature aggregation could dynamically regulate language contributions, mitigating such effects in practice, as evidenced by the reduced sensitivity to misleading descriptions (Table~\ref{table:modality_bias_task_formulation}). 
Fully addressing language bias remains an open direction.

\section*{Ethics Statement}
This work builds upon publicly available video datasets and does not involve private or personally identifiable data. However, vision-language models may inherit biases from pre-training corpora and textual supervision, which can propagate into temporal action localization. If misused in surveillance or monitoring settings, action localization systems could enable harmful profiling or intrusive behavioral analysis. In addition, biased descriptions or uneven data coverage may affect some activities, populations, or environments disproportionately when such systems are deployed in the real world. To mitigate these risks, we recommend dataset auditing, bias evaluation across activity and environment groups, and uncertainty reporting before any deployment. Our method is intended for academic research on robust and fair video understanding rather than high-stakes decision-making.

\bibliography{custom}

\clearpage
\appendix

\section{Supplementary Material}

\begin{figure*}[b] 
    \centering
    \begin{adjustbox}{width=1\linewidth,center}
\includegraphics{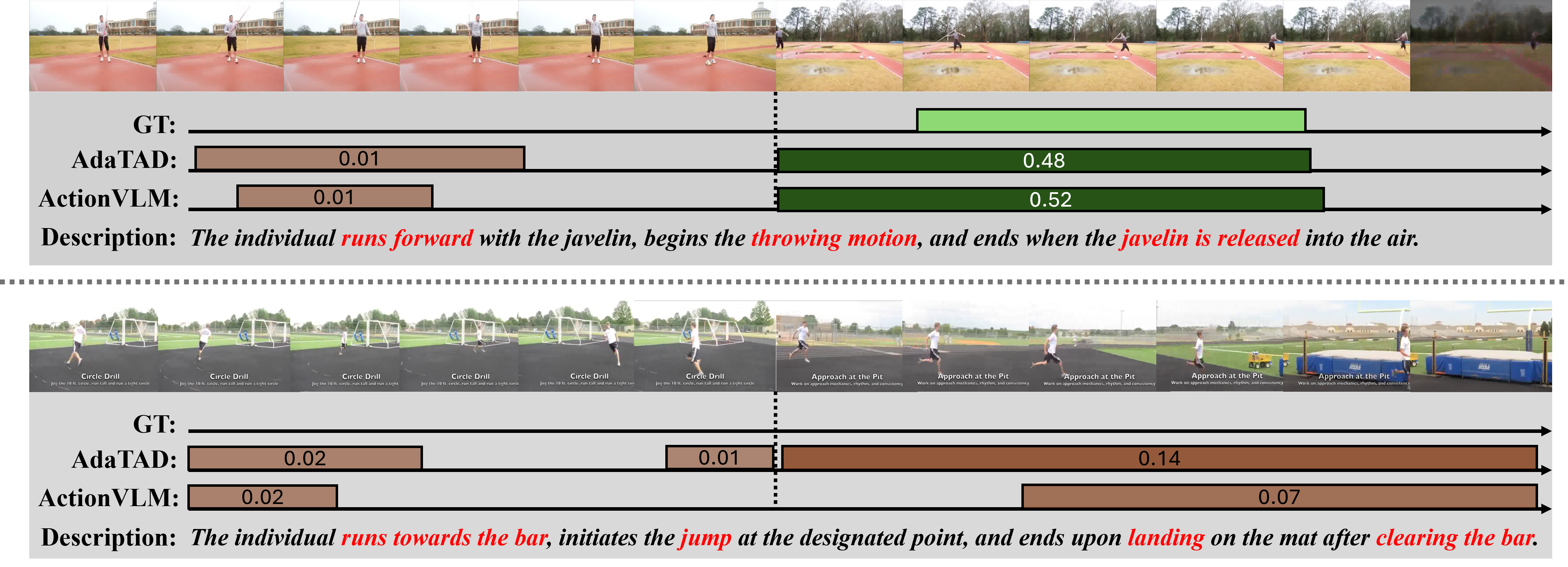} 
\end{adjustbox}
\caption{
Qualitative visualization under visual ambiguity.
From top to bottom: (1) video frames, (2) ground truth (GT), (3) predicted boundaries with confidence (classification score from detection head), and (4) corresponding action descriptions.
True positives are shown in \textcolor{Green4}{green}, while visually similar false positives without GT are highlighted in \textcolor{Chocolate4}{brown}.
Lower confidence and shorter spans indicate better calibration on ambiguous clips.
} 
\label{Fig:visualisation} 
\end{figure*}

\subsection{Analysis on Visual Ambiguity}
\label{appx:visualization_ambiguity}

\textbf{Quantitative Analysis.}
To evaluate whether ActionVLM can disambiguate visually similar actions, we design a dedicated benchmark on THUMOS14.
Since existing datasets rarely capture such ambiguity, we manually collect 40 challenging ambiguous clips—two for each of the 20 categories—where subjects perform incomplete or misleading poses.
Each full video is processed by the model, while evaluation is restricted to these ambiguous segments.
As multiple temporal proposals are generated per video (Sec.~\ref{sec:rpn}), we retain only the highest-confidence one for clarity.

In ambiguous settings, a reliable model should reflect uncertainty—either by reducing its prediction confidence or by refining its temporal boundaries.
We therefore introduce two complementary metrics: the mean predicted confidence ($mConf$) and the mean normalized span length ($mLen$), which jointly measure how cautiously and precisely the model localizes actions.
Furthermore, we report accuracies at different $mLen$ thresholds, defined as the proportion of predictions with span lengths below the threshold.

As shown in Table~\ref{table:vs_quantitative}, incorporating language priors consistently lowers both $mConf$ and $mLen$, indicating better uncertainty calibration and temporal reasoning.
Compared to the vision-only baseline ($\lambda_{lang}$=0.0), our learnable modality weight reduces confidence by 1.2\% and span length by 8.1\%, confirming that adaptive language integration helps the model stay grounded under visual ambiguity.

\begin{table}[t]
\centering
\begin{adjustbox}{width=0.52\textwidth,center}
\begin{tabular}{cccccc}
\hline
                     &  \multicolumn{5}{c}{THUMOS14}  \\ 
$\lambda_{lang}$     &$mConf\downarrow$&$mLen\downarrow$   & $acc@0.3\uparrow$& $acc@0.5\uparrow$ & $acc@0.7\uparrow$ \\
\hline
1.0                  &   4.7            &        52.5       &       12.5       &        37.5      &         67.5        \\

0.8                  &   4.6            &        52.4       &       12.5       &        37.5       &        70.0        \\

0.6                  &   4.5            &        51.2       &       15.0       &        42.5       &        72.5        \\

0.4                  &   4.4            &        50.0       &       15.0       &        42.5       &        75.0    \\

0.2                  &   4.6            &        51.9       &       10.0       &       35.0       &         67.5       \\

0.0                  &   5.0            &        56.9       &       5.0        &        30.0       &        57.5        \\

\textbf{learnable}   &   \textbf{3.8}   &     \textbf{48.8} &    \textbf{15.0} &     \textbf{45.0} &      \textbf{80.0}          \\
\hline
\end{tabular}
\end{adjustbox}
\caption{Quantitative results (all in \%) on visually similar clips under different language feature weights.}
\label{table:vs_quantitative}
\end{table}

\textbf{Qualitative Analysis.}
To further assess ActionVLM’s behavior under visual ambiguity, we visualize three distractors and one true instance from the javelin throw and high jump categories.
The distractors exhibit strong visual resemblance to the target actions—e.g., a subject holding a javelin without throwing, or running toward the high-jump bar without actually jumping—creating challenging false positives for vision-only models.

As shown in Fig.~\ref{Fig:visualisation}, models without language guidance tend to overfit to low-level motion cues and misclassify incomplete actions as valid ones.
In contrast, ActionVLM leverages temporal semantics encoded in language descriptions to infer action completeness, suppressing false positives and tightening temporal spans.
This yields notably lower confidence and shorter predictions on distractors, while maintaining strong activation for true actions.
In the most challenging case (the second case of high jump), ActionVLM reduces span length by 34\% and confidence by 50\%, demonstrating its capacity to reason over action intent and completion, rather than surface-level appearance.
                                    
\begin{table*}[t]
\centering
\begin{adjustbox}{width=0.90\textwidth,center}
\begin{tabular}{clcccc}
\hline
                 & \textbf{Model}        &\textbf{Window Size} &\textbf{Training Latency (ms)}&\textbf{Evaluation Latency (ms)}&\textbf{mAP}\\
\hline
\multirow{2}{*}{(a)}&VTimeLLM \cite{huang2024vtimellm}          & 128         &   213       &   928         & 33.4 \\
                    &Timemarker \cite{chen2024timemarker}       & 128         &   229       &   1079        & 42.9 \\

\hline
\multirow{2}{*}{(b)}&AdaTAD-S \cite{liu2024end}                 & 768         &   265       &   306       & 68.8  \\
                    &AdaTAD-B \cite{liu2024end}                 & 768         &   601       &   626       & 71.5    \\
\hline
\multirow{2}{*}{(c)}&\textbf{ActionVLM-S-3B (ours)}             & 768         &   289       &   374       & 72.0    \\
                    &\textbf{ActionVLM-B-3B (ours)}             & 768         &   656       &   741       & 74.2        \\
\hline
\end{tabular}
\end{adjustbox}
\caption{Comparison with state-of-the-art methods in terms of efficiency on THUMOS14. (a) Conventional VLMs. (b) Vision-only Baselines. (c) Our proposed ActionVLMs. }
\label{table:efficiency}
\end{table*}

\begin{table*}[t]
\centering
\begin{adjustbox}{width=0.75\textwidth,center}
\begin{tabular}{lll|cccccc}
\hline
& & &\multicolumn{6}{c}{\textbf{Epic-Kitchens 100}}\\
\textbf{Model}                           &\textbf{Vision}&\textbf{E2E}&\textbf{0.1} &\textbf{0.2} &\textbf{0.3} &\textbf{0.4} &\textbf{0.5} &\textbf{Avg.} \\
\hline
\textit{Verb Task}&&&&&&\\
ActionFormer \cite{zhang2022actionformer}& SlowFast      &            & 27.7         & 26.8         & 25.6        & 24.4          & 20.5         & 24.9   \\
ActionFormer \cite{zhang2022actionformer}& VideoMAEv1-L  &            & 32.7         & 31.6         & 29.1        & 26.7          & 23.6         & 28.7   \\
AdaTAD  \cite{liu2024end}                & VideoMAEv1-L  &\chk        & 33.1         & 32.2         & 30.4        & 27.5          & 23.1         & 29.3   \\
\textbf{ActionVLM-L-3B (ours) }          & VideoMAEv1-L  &\chk        & \textbf{34.0}& \textbf{33.3}& \textbf{31.4}& \textbf{28.8}&\textbf{24.8} &\textbf{30.5}\\
\hline
\textit{Noun Task}&&&&&&\\
ActionFormer \cite{zhang2022actionformer}& SlowFast      &            & 25.8         & 24.7         & 22.8        & 20.8          & 17.5         & 22.3   \\
ActionFormer \cite{zhang2022actionformer}& VideoMAEv1-L  &            & 31.3         & 29.7         & 27.2        & 25.3          & 21.3         & 26.9   \\
AdaTAD  \cite{liu2024end}                & VideoMAEv1-L  &\chk        & 32.4         & 31.6         & 30.1        & 27.4          & 24.6         & 29.3   \\
\textbf{ActionVLM-L-3B (ours) }          & VideoMAEv1-L  &\chk        &\textbf{32.8} & \textbf{32.1}&\textbf{30.7}& \textbf{28.2} &\textbf{26.1} &\textbf{30.0}\\
\hline
\end{tabular}
\end{adjustbox}
\caption{Results of mAPs (\%) at different IoU thresholds on Epic-Kitchens 100. Best results are highlighted in \textbf{bold}.}
\label{table:benchmark_epic}
\end{table*}

\subsection{Analysis on Efficiency}

Although incorporating VLMs can improve the reliability of reasoning \cite{ye2026generating,ye2025knowledge}, it inevitably incurs additional inference overhead.
Table~\ref{table:efficiency} compares ActionVLM with both conventional VLMs and strong vision-only baselines in terms of computational efficiency and accuracy on THUMOS14.
We report average training and inference latency per video clip.
The window size denotes the number of frames processed in a single temporal chunk, which affects both efficiency and temporal context coverage.
We set a smaller window size for conventional VLMs to fit GPU memory, while ActionVLM and AdaTAD use longer windows to capture richer temporal dynamics following the original setting \cite{liu2024end}.

As shown in Table~\ref{table:efficiency}, the heavy computation of conventional VLMs severely limits their scalability to downstream tasks.
VTimeLLM and Timemarker process short clips with 128 frames yet still exhibit high inference latency due to dense patch tokens and autoregressive decoding.
This inefficiency makes direct deployment to long-sequence temporal action localization (TAL) impractical.

To make VLMs viable for video tasks, we introduce two acceleration strategies.
First, instead of representing each frame by hundreds of spatial patch tokens, we apply a mean-pooling layer over spatial dimensions, compressing them into a single visual token.
This reduces token length by more than one order of magnitude without compromising temporal fidelity, enabling efficient long-sequence modeling.
Second, we replace the conventional token-by-token decoding for textual proposal generation (Eq.~\ref{eq:loss_text}) with a lightweight auxiliary generation scheme.
During training, we employ the conventional practice of teacher-forcing strategy \cite{williams1989teacherforcing}, which enables parallel generation and has little effect on the efficiency, while at inference we restrict the output $\hat{T}_{pred}$ to one token since we rely on feature aggregation while regarding text generation as an auxiliary task.
This preserves semantic supervision with negligible computational cost.

As shown in Table~\ref{table:efficiency}, these designs substantially improve efficiency.
Compared with conventional VLMs, ActionVLM reduces inference latency by more than two times while processing six times longer sequences.
When compared with the strong vision-only baseline AdaTAD, ActionVLM introduces only a marginal overhead yet achieves consistent accuracy gains of up to 3.2\% mAP.
This demonstrates that multimodal reasoning can be achieved efficiently, bridging the gap between language-augmented understanding and practical large-scale TAL deployment.

\begin{table*}[t]
\centering
\begin{adjustbox}{width=0.90\textwidth,center}
\begin{tabular}{ll|cccccc}
\hline
& & \multicolumn{6}{c}{\textbf{THUMOS14}}\\
\textbf{Model}&\textbf{Vision} &\textbf{0.3}  &\textbf{0.4}  &  \textbf{0.5}&  \textbf{0.6}&  \textbf{0.7}&\textbf{Avg.} \\
\hline
AdaTAD  \cite{liu2024end}  & VideoMAEv1-B  \cite{tong2022videomae}            & 87.0          & 82.4          & 75.3         & 63.8          & 49.2         & 71.5    \\
AdaTAD  \cite{liu2024end}  & VideoMAEv2-B   \cite{wang2023videomae}           & 87.5          & 83.1          & 76.0         & 66.6          & 51.3         & 72.8    \\
\hline
ActionVLM-3B (ours)        & InternViT-300M-V2.5 \cite{chen2024internvl}      & 72.6         & 67.5         & 59.3         & 48.4         & 35.4         & 56.7\\
ActionVLM-B-3B (ours)      & VideoMAEv1-B   \cite{tong2022videomae}           & 88.6         & 84.2         & 76.1         & 64.4         & 49.6         & 72.6\\
ActionVLM-B-3B (ours)      & InternVideo2-B \cite{wang2024internvideo2}       & \textbf{88.8}& 84.5         & 77.3         & 66.5         & 51.1         & 73.6   \\
\textbf{ActionVLM-B-3B (ours)}& VideoMAEv2-B \cite{wang2023videomae}          &\textbf{88.8} &\textbf{85.1} &\textbf{78.2} &\textbf{67.1} &\textbf{51.8} &\textbf{74.2}\\
\hline
\end{tabular}
\end{adjustbox}
\caption{Ablation on different vision encoders.}
\label{table:vision_ablation}
\end{table*}

\begin{table*}[t]
\centering
\begin{adjustbox}{width=1\textwidth,center}
\begin{tabular}{clcccc|cccccc}
\hline
                    &                                                                &                &             &             &             & \multicolumn{6}{c}{\textbf{THUMOS14}}\\
                 & \textbf{Model}&\textbf{$T_{desc}$} &\textbf{[cls]}&\textbf{[loc]}&\textbf{$\mathcal{L}_{tg}$}&\textbf{0.3}  &\textbf{0.4}  &  \textbf{0.5}&  \textbf{0.6}&  \textbf{0.7}&\textbf{Avg.} \\
\hline
\multirow{2}{*}{(a)}&ActionVLM w/o action description $T_{desc}$                          &     &\chk &\chk &\chk &  85.9        &  81.4        &  74.1        &   63.8       &   48.2       &    70.7 \\
                    &\textbf{ActionVLM} (full, with priors in description)           &\chk &\chk &\chk &\chk &\textbf{87.2}&\textbf{82.9}&\textbf{75.8}&\textbf{64.6}&\textbf{49.6}&\textbf{72.0} \\

\hline
\multirow{3}{*}{(b)}&ActionVLM w/o special tokens (extract language feature from vision tokens)                &\chk &     &     &\chk &  85.8        &  81.9        &  74.6  &   63.5       &   48.2       &    70.8  \\
                    &ActionVLM w/o [loc] tokens (share a single special token)                    &\chk &\chk &   &\chk &  86.6        &  82.4        &   75.2       &   64.2       &    49.4      &    71.6    \\
                    &\textbf{ActionVLM} (full, with both [cls] and [loc])            &\chk &\chk &\chk &\chk &\textbf{87.2}&\textbf{82.9}&\textbf{75.8}&\textbf{64.6}&\textbf{49.6}&\textbf{72.0} \\
\hline
\multirow{2}{*}{(c)}&ActionVLM (w/o text generation loss $\mathcal{L}_{tg}$)         &\chk &\chk &\chk &     &  86.0        &  81.9        &  74.8        &   64.0       &   49.3       &    71.2    \\
                    &\textbf{ActionVLM} (full, with $\mathcal{L}_{tg}$)              &\chk &\chk &\chk &\chk &\textbf{87.2}&\textbf{82.9}&\textbf{75.8}&\textbf{64.6}&\textbf{49.6}&\textbf{72.0} \\
\hline
\end{tabular}
\end{adjustbox}
\caption{Ablation on (a) action descriptions, (b) feature tokens, and (c) text generation loss $\mathcal{L}_{tg}$.
}
\label{table:feat_aggregation}
\end{table*}

\subsection{Additional Experiments on EPIC-Kitchens}
To validate the generalizability of our model on diverse datasets, we further conduct evaluation on EPIC-Kitchens-100 \cite{damen2020epic}. 
EPIC-Kitchens-100 is a large-scale egocentric benchmark comprising 700 videos (about 100 hours) recorded in 45 kitchens, with fine-grained verb and noun annotations. 
Its domain specificity and egocentric action patterns differ substantially from the visual and linguistic distributions commonly seen in large-scale VLM pretraining corpora, making high absolute mAP particularly challenging \cite{liu2024end}. 

We use GPT-4o-mini to generate short descriptions for each class used in our EPIC-Kitchens-100 evaluation setup. This results in 1,283 description tokens for the 78 verb classes and 2,277 tokens for the 211 noun classes. 
Importantly, our runtime is dominated by video encoding rather than language modeling, so the text token budget is not the computational bottleneck. 

Despite this increased difficulty, ActionVLM-L-3B consistently outperforms AdaTAD across both verb and noun tasks (Table~\ref{table:benchmark_epic}), with particularly notable gains on verbs of 1.2\% average mAP. 
This trend is expected: verbs describe actions and motion primitives, whereas language priors provide meaningful semantic constraints that help disambiguate visually subtle or incomplete cues in egocentric footage. 
In contrast, noun predictions depend primarily on object appearance, where linguistic context contributes less discriminative information, resulting in smaller—but still positive—improvements.
These results highlight that ActionVLM not only generalizes beyond standard third-person TAL datasets but also benefits most from scenarios where language provides actionable semantic structure for interpreting complex visual dynamics.

\subsection{Additional Ablation on Vision Encoder}
Table \ref{table:vision_ablation} shows that directly adopting the raw vision encoder from InternVL3 (InternViT-300M-V2.5) \cite{chen2024internvl} yields substantially weaker TAL performance compared with VideoMAE-based \cite{tong2022videomae} variants. 
This degradation is largely attributable to two factors: (i) the InternVL3 vision encoder is not pretrained on video-centric action datasets such as Kinetics \cite{kay2017kinetics}, and (ii) it processes single frames rather than 16-frame snippets, preventing it from modeling motion dynamics and temporal continuity. 
By contrast, VideoMAE leverages masked video reconstruction over multi-frame inputs, enabling stronger temporal representation learning that better aligns with the requirements of TAL. 
As a result, replacing the InternVL3 encoder with VideoMAE leads to large gains across all IoU thresholds. 
Moreover, even under the same vision backbone (VideoMAEv2-B), our ActionVLM-B-3B still surpasses AdaTAD by 1.4\% on average mAP, confirming the effectiveness of our overall design.

\subsection{Additional Ablation on Language Feature Generation}
\textbf{Action Descriptions.}
As shown in Table~\ref{table:feat_aggregation}(a), removing descriptions leads to a 1.2\% mAP drop, confirming their value as auxiliary semantic priors. 
However, because these descriptions are not visually grounded, mismatches with vision content may introduce bias. 
Our proposed reweighting mitigates such mismatches, ensuring language cues remain complementary rather than misleading.

\textbf{Special Tokens.}
As shown in Table~\ref{table:feat_aggregation}(b), assigning separate tokens for classification and localization disentangles the objectives and enables specialized representations, yielding superior performance compared to shared or vision tokens.

\textbf{Text Generation.}
Table~\ref{table:feat_aggregation}(c) shows that the auxiliary text generation task strengthens vision-language alignment while providing additional supervision. 
Removing it weakens the shared semantic space and leads to a 0.7\% mAP drop.

\subsection{Histogram of Linguistically Preferred Boundaries}
\label{appx:hist}
\begin{figure}[H] 
    \centering
    \begin{adjustbox}{width=1\linewidth,center}
\includegraphics{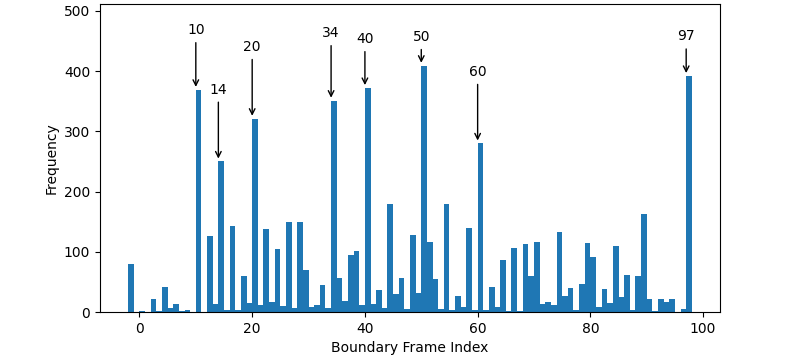} 
\end{adjustbox}
\caption{Histogram of frequently predicted temporal boundaries. The eight most frequent frame indices are highlighted with arrows.} 
\label{Fig:hist} 
\end{figure}

As shown in Fig.~\ref{Fig:modality_bias} and discussed in Sec.~\ref{sec:exp:modality_bias}, QA-based VLMs exhibit a tendency to predict linguistically preferred temporal boundaries, such as coarse or rounded timestamps.
To verify that this behavior is systematic rather than anecdotal, we further analyze the distribution of predicted boundaries on our dataset.
Specifically, we construct a histogram of boundary predictions produced by TimeMarker using a fixed window size of 100 frames. 
For each localization instance, the model receives 100 frames and outputs predicted start and end frame indices, which are aggregated across all samples (Fig.~\ref{Fig:modality_bias} shows the aggregated indices, while Fig.~\ref{Fig:hist} presents the raw ones).

As illustrated in Fig.~\ref{Fig:hist}, two clear patterns emerge: round-number peaks at multiples of ten (e.g., 10, 40, 60) and dataset-specific peaks aligned with fine-tuning data statistics (e.g., 14, 34, 97). These patterns provide further evidence that the predicted boundaries are influenced by linguistic priors rather than purely visual evidence.

\subsection{Model Sensitivity to Language Granularity}

Action descriptions improve TAL by providing auxiliary semantic priors, but they also introduce an additional text-token budget. 
We further ask a practical question: how much description detail is actually needed? 
In particular, we study whether the full sentence descriptions used in ActionVLM can be replaced by cheaper alternatives without sacrificing accuracy.

We compare two budget-saving variants against the full setting. 
The first is \emph{keywords-only}, where each class is represented by only three words summarizing the start, middle, and end motions. 
The second is \emph{merged descriptions}, where semantically similar actions (e.g., LongJump \& HighJump) share one coherent sentence. Table~\ref{table:language_granularity} reports the results.

\begin{table}[t]
\centering
\small
\begin{tabular}{lcc}
\hline
Model & \#Token & mAP \\
\hline
w/o description & 0   & 70.7 \\
keywords-only & 219 & 71.0 \\
merged descriptions & 659 & 71.9 \\
ActionVLM-S-3B (full) & 742 & 72.0 \\
\hline
\end{tabular}
\caption{Sensitivity to language granularity on THUMOS14.}
\label{table:language_granularity}
\end{table}

While keywords-only descriptions reduce the text budget substantially, they provide only marginal gains over removing descriptions, whereas merged descriptions nearly match the full model. 
This indicates that preserving coherent action semantics is more important than retaining every class-specific phrase. 
Since our efficiency analysis in Table \ref{table:efficiency} shows that our runtime is dominated by video encoding rather than language modeling, we use the full descriptions by default to achieve the best performance.

\subsection{Analysis on Performance Gain}
\label{appx:gain}
In Temporal Action Localization (TAL), even seemingly small improvements in mAP often correspond to substantial underlying advances.

\textbf{mAP Sensitivity in TAL.}
TAL is a long-tailed and highly imbalanced task, where many frequent and visually distinctive actions are already well localized by vision-only models. 
Consequently, mAP—averaged across classes and strict tIoU thresholds—is dominated by these easy cases, making further improvements inherently difficult. 
As a result, even small increases in overall mAP often correspond to substantial gains on rare or visually ambiguous actions. 
As shown in Table~\ref{table:modality_bias_difficulty} and Table~\ref{table:lang_weight}, ActionVLM significantly improves AP on visually ambiguous actions, indicating that the observed gains are concentrated on the most challenging cases where modality bias is most harmful.

\textbf{Annotator Subjectivity and Semantic Alignment.}
Both THUMOS14 and ActivityNet-1.3 exhibit notable annotator subjectivity, particularly for long or semantically complex actions with ambiguous boundaries. 
In such cases, annotators implicitly rely on high-level semantic understanding rather than low-level visual cues alone. 
Language models naturally encode such semantic priors and thus better align with human annotation tendencies. 
By selectively incorporating language cues through debiasing aggregation, ActionVLM improves robustness to boundary ambiguity without overriding visual evidence, leading to more annotation-consistent predictions.

\textbf{Backbone Saturation.}
Recent advances in TAL have largely been driven by increasingly powerful video backbones \cite{liu2024end, yang2024adapting}, which are now approaching performance saturation. 
With strong temporal encoders already in place, large mAP jumps are no longer realistic. 
In this regime, consistent improvements across datasets and IoU thresholds signify the correction of errors that cannot be resolved through better visual representations alone. 
ActionVLM addresses this limitation by mitigating modality bias, offering a complementary axis of improvement beyond backbone scaling.

\subsection{Action Categories by Visual Ambiguity}
\label{appx:action_category}
THUMOS14 contains 20 action categories.
For the analysis in Sec.~\ref{sec:exp:language_advantage}, we group these actions into three subsets based on their visual ambiguity, measured by the performance of a vision-only baseline.

Specifically, actions with high vision-only mAP are considered visually unambiguous, as they can be reliably localized without language cues, while low-performing actions exhibit higher visual ambiguity.

Based on this criterion, we define:
\begin{itemize}[itemsep=0pt,topsep=0pt,parsep=0pt,leftmargin=10pt]
\item \textbf{Hard}: Billiards, FrisbeeCatch;
\item \textbf{Easy}: ThrowDiscus, LongJump, HighJump, HammerThrow;
\item \textbf{Medium}: the remaining actions.
\end{itemize}

\section{Prompts}
\subsection{The Prompts for Action Descriptions Generation}
\label{appx:prompt_descriptions}

We introduce action descriptions $T_{desc}$ as prior knowledge for localization. These are generated by prompting GPT-4o \cite{gpt4o} to produce detailed textual depictions of representative start and end poses for each action category, without access to visual input. The prompts used for generation are as follows:

\vspace{5pt}

\emph{
"Describe the action from the start pose to the end pose:
e.g. BasketballDunk: The individual starts to jump towards the hoop while holding the basketball and ends when the ball is forcefully pushed through the basket."
}

\vspace{5pt}

The full prompts for each action are recorded in our code. 
Here we provide the generated results used in our paper as examples:

\vspace{5pt}

\emph{
BaseballPitch: The individual starts by preparing and winding up with the baseball in hand, then initiates the pitching motion, and ends when the ball is released and thrown toward the batter.
}

\emph{
Billiards: The individual starts by positioning the cue stick and aiming at the cue ball, then strikes the cue ball with the stick, and ends when the balls come to a complete stop after the shot.
}

\emph{
CleanAndJerk: The individual starts by gripping the barbell on the ground, then lifts it to the shoulders in the "clean" phase, pauses briefly, and continues by explosively lifting it overhead in the "jerk" phase, ending when the barbell is held steady overhead with full control.
}

\emph{
CliffDiving: The individual starts by standing at the edge of a cliff or platform, then jumps off and performs acrobatic movements while descending, ending when they enter the water below.
}

\emph{
CricketBowling: The individual starts by taking a run-up toward the wicket, then releases the ball with a straight arm towards the batsman, ending when the ball reaches the batsman or wicketkeeper.
}

\emph{
CricketShot: The individual starts by preparing the bat stance as the ball approaches, then swings the bat to hit the ball, ending when the ball is struck and sent away from the batsman.
}

\emph{
Diving: The individual starts by running or standing at the edge of a diving board or platform, then jumps and performs a controlled descent into the water, ending upon water entry.
}

\emph{
FrisbeeCatch: The individual starts by tracking the flying frisbee, positions their hands or body to intercept, and ends when they successfully grasp or trap the frisbee.
}

\emph{
GolfSwing: The individual begins by addressing the golf ball, then swings the club in a controlled arc to strike the ball, ending when the ball is launched toward the target.
}

\emph{
HammerThrow: The individual starts by gripping the hammer, initiates spinning rotations to build momentum, and ends when they release the hammer into the throwing sector.
}

\emph{
HighJump: The individual runs towards the bar, initiates the jump at the designated point, and ends upon landing on the mat after clearing the bar.
}

\emph{
JavelinThrow: The individual runs forward with the javelin, begins the throwing motion, and ends when the javelin is released into the air.
}

\emph{
LongJump: The individual begins running along the track, then takes off from the takeoff board, and ends when they land in the sandpit.
}

\emph{
PoleVault: The individual starts by sprinting down the runway holding the pole, plants the pole into the vault box, uses it to propel upward over the bar, and ends when they land safely on the mat.
}

\emph{
Shotput: The individual begins by positioning the shot near the neck, then uses a pushing motion to launch the shot forward, ending when the shot lands on the ground.
}

\emph{
SoccerPenalty: The individual starts by approaching the ball placed at the penalty mark and ends after striking the ball towards the goal.
}

\emph{
TennisSwing: The individual starts by preparing their stance and tracking the incoming ball, then swings the racket to strike the ball, ending when the ball is hit and sent back over the net.
}

\emph{
ThrowDiscus: The individual starts by gripping the discus, performs a spinning motion to gain momentum, and ends when the discus is released into the air.
}

\emph{
VolleyballSpiking: The individual begins by jumping near the net, swings their arm forcefully to hit the ball downward over the net, and ends when the ball crosses into the opponent’s court.
}

\vspace{5pt}

\subsection{Instruction Prompt for Textual Action Proposal Generation}
\label{appx:prompt_generation}

We employ an auxiliary textual proposal generation task to better align the vision and language modalities of the VLM.  
The instruction prompt $T_{\mathrm{ins}}$ guides the model to perform temporal action localization (TAL) in a text-to-proposal manner, transforming visual observations into ranked textual predictions based on confidence scores.  
This design encourages the model to express its temporal reasoning explicitly in natural language, facilitating cross-modal consistency between detection and generation.

\vspace{5pt}

\emph{
"This is a video reasoning for the action localization task. You should identify the start and end frames of each action in the given video and output the ranked list of action proposals according to their confidence.  
The candidate actions include: $\langle T_{\mathrm{desc}}\rangle$""
}

\vspace{5pt}

\end{document}